\documentclass[letterpaper, 10 pt, conference]{ieeeconf}  

\IEEEoverridecommandlockouts                              

\title{\LARGE \bf
A Lightweight Crowd Model for Robot Social Navigation
}

\author{Maryam Kazemi Eskeri$^{1}$, Thomas Wiedemann$^{2,}$$^{3}$, Ville Kyrki$^{1}$, Dominik Baumann$^{1}$, and Tomasz Piotr Kucner$^{1}$
\thanks{This work was supported by the Research Council of Finland Flagship programme: Finnish Center for Artificial Intelligence FCAI. We acknowledge the computational resources provided by the Aalto Science-IT project and CSC, Finnish IT Center for Science.}
\thanks{$^{1}$Department of Electrical Engineering and Automation, Aalto University, 02150 Espoo, Finland (e-mail: firstname.lastname@aalto.fi)
$^{2}$Chair of Perception for Intelligent Systems, TU Munich, Germany (e-mail: firstname.lastname@tum.de)
$^{3}$Institute of Communications and Navigation, German Aerospace Center (DLR), Germany (e-mail: firstname.lastname@dlr.de)
}
}%

\usepackage{amsmath}  
\usepackage{amssymb}
\usepackage{mathrsfs} 

\usepackage{cleveref}
\usepackage{booktabs}
\usepackage{multirow}
\usepackage{graphicx}

\usepackage[table]{xcolor}
\usepackage{pgf}
\usepackage{pgfplots} 
\usepackage{siunitx}
\usepackage[acronym]{glossaries}
\usepackage{mathtools}
\usepackage{everypage}
\usepackage{stfloats} 
\usepackage{float}
\usepackage[final]{changes}
\usepackage{multicol}
\usepackage{lipsum}
\usepackage{mwe}

\usepackage{graphicx} 

\usepackage{fancyhdr}
\newcommand{\mytitle}{\textbf{Accepted final version.}
To appear in \textit{Proceedings of the IEEE}.\\
\copyright 2025 IEEE. Personal use of this material is permitted. Permission
from IEEE must be obtained for all other uses, in any current or future
media, including reprinting/republishing this material for advertising or
promotional purposes, creating new collective works, for resale or
redistribution to servers or lists, or reuse of any copyrighted component of
this work in other works.}
\begin{document}

\maketitle

\thispagestyle{fancy}
\pagestyle{plain}
\newacronym{lstm}{LSTM}{Long-Short Term Memory network}
\newacronym{gan}{GAN}{Generative Adversarial Network}
\newacronym{convrnn}{ConvRNN}{Convolutional Recurrent Neural Network}
\newacronym{mae}{MAE}{Mean Absolute Error}

\newtheorem{problem}{Problem}
\begin{abstract}
Robots operating in human-populated environments must navigate safely and efficiently while minimizing social disruption. Achieving this requires estimating crowd movement to avoid congested areas in real-time. Traditional microscopic models struggle to scale in dense crowds due to high computational cost, while existing macroscopic crowd prediction models tend to be either overly simplistic or computationally intensive. In this work, we propose a lightweight, real-time macroscopic crowd prediction model tailored for human motion, which balances prediction accuracy and computational efficiency. Our approach simplifies both spatial and temporal processing based on the inherent characteristics of pedestrian flow, enabling robust generalization without the overhead of complex architectures.  We demonstrate a 3.6 times reduction in inference time, while improving prediction accuracy by \SI{3.1}{\percent}. Integrated into a socially aware planning framework, the model enables efficient and socially compliant robot navigation in dynamic environments. This work highlights that efficient human crowd modeling enables robots to navigate dense environments without costly computations.

\end{abstract}

\section{INTRODUCTION}

Nowadays, robots are increasingly deployed in human-centered environments e.g. airports~\cite{triebel2016spencer}, or shopping malls~\cite{sabelli2016robovie}.
In all of these applications, for robots to safely co-exist with humans in dynamic settings, it is essential to navigate in a socially compliant manner. Efficient navigation in human-shared environments goes beyond static obstacle avoidance, but it involves navigating without being intrusive or disruptive to people's movements and interactions. Achieving this requires the ability to understand and predict general motion patterns
and adapt robot paths in real-time.

Over the past decades, many robot motion planning approaches have been developed to navigate around dynamic obstacles in real-time~\cite{singamaneni2024survey,mavrogiannis2023core}, such as the Dynamic Window Approach (DWA)~\cite{molina2019go,liu2025improved}. However, the computational demands on the local planner increase significantly with a growing number of humans, which hinders navigation efficiency and safety and can lead to longer travel times or even collision with pedestrians. 
Therefore through enabling global planners to generate pattern-aware trajectories addresses the limitations of currently existing approaches.

Crowd behavior can be modeled using both microscopic and macroscopic approaches. Microscopic modeling focuses on the actions of individual pedestrians and their interactions. Social force models (SFMs)~\cite{mehran2009abnormal} are one of the most fundamental techniques in this category. SFMs treat pedestrians like particles influenced by psychological forces that motivate them to reach their goals while 
keeping a distance from other agents. Although SFMs effectively capture important crowd behaviors, they depend on manually-tuned energy potential functions. This can lead to a 
suboptimal behavior and cause safety issues in robot navigation~\cite{cheng2024multi}. Another major challenge in this context is the freezing robot problem~\cite{trautman2010unfreezing}, which occurs when robots become stuck due to uncertainty in dense, dynamic environments.

Recent advances in deep learning have shifted the paradigm toward data-driven crowd prediction. Models like the Social \acrfull{lstm}~\cite{alahi2016social} and the Social \acrfull{gan}~\cite{gupta2018social} effectively capture the spatial and temporal dependencies in pedestrian movements, showing better performance than traditional hand-crafted methods. Over time, a wide range of variations of \acrshort{lstm} and \acrshort{gan} architectures have been created to improve predictive accuracy and better suit different environments~\cite{sadeghian2019sophie,vemula2018social,huang2025interaction}. Although these neural network methods have shown significant success in forecasting human movements in low-density scenarios, their applicability to dense crowds remains an open research question.

Microscopic crowd prediction models focus on predicting individual behavior rooted in distinct goals~\cite{treuille2006continuum} of each individual.
Such solutions are not only intractable for densely populated environments, but also as the density of the crowd increases, individual behavior becomes more constrained, reducing the influence of microscopic properties on the overall behavior of the crowd~\cite{narain2009aggregate}. 
This makes macroscopic modeling particularly 
interesting 
as 
such methods
look at the overall behavior of the crowd
making them more suitable for understanding and predicting movement in dense crowds.

Macroscopic approaches model crowds as continuous spatial fields of density and flow velocity. ~\cite{treuille2006continuum} frame crowd motion as continuum dynamics, whereas~\cite{narain2009aggregate} link it to fluid dynamics. This perspective enabled the efficient simulation of large groups. However, these approaches rely on static or deterministic models of crowd behavior, which restricts their relevance to real-world situations where human motion is non-deterministic and influenced by dynamic social context. 

A state-of-the-art approach in crowd modeling is presented by Kiss et al. ~\cite{kiss2021probabilistic}, where macroscopic crowd behavior is modeled using the \acrfull{convrnn}, a deep recurrent architecture initially developed for precipitation forecasting \cite{shi2015convolutional}. While effective in capturing spatiotemporal patterns, this model is optimized for complex, high-dimensional weather data, which differ significantly from structured and often more predictable human behavior. As a result, it risks overfitting when applied to crowd scenarios. 
Finally, the model’s high computational and training costs limit its ability to perform fast, adaptive forecasting in dynamic environments.

In conclusion the limitations of existing micro- and macroscopic approaches %
highlight the need for more lightweight, flexible approaches to crowd behavior modeling. In dynamic environments, where conditions change rapidly, it is crucial for models to provide timely and accurate predictions. Current methods often sacrifice computational efficiency for high accuracy, making them less suitable for real-time applications. Thus, to address these needs and limitations, in this work, we focus on optimizing the model to perform effectively in real-world scenarios. 
At the same time, the proposed solution ensures that it can quickly adjust to evolving crowd dynamics while maintaining a balance between computational efficiency and prediction accuracy. To demonstrate the effectiveness of the proposed lightweight model, we evaluate it on a real-world crowd dataset and compare the robot's planned trajectories when using our lightweight predictor versus a high-dimensional baseline~\cite{kiss2021probabilistic}. Our results show that the lightweight model achieves comparable navigation performance while offering significantly improved computational efficiency, reducing inference time 3.6 times,  making it well-suited for dynamic, real-world applications.

\section{Problem formulation}
\label{problemformulation}
Let us consider a mobile robot navigating through an environment populated by pedestrians, where crowd evolution needs to be predicted and accommodated for real-time planning. Let $\mathcal{C} \subseteq \mathbb{R}^2$ represent the environment in which humans move, discretized into a grid of size $H \times W$. At each time step $t \in \mathbb{N}$, the global environment state is represented by $X[t]\subseteq \mathbb{R}^{H \times W\times d}$, where $d$ is the number of macroscopic crowd features encoded per cell. These features include spatial distributions of pedestrian density and flow characteristics such as average velocity and variance, which are described in detail in~\cref{crowd representation}.

\begin{problem}
Given a sequence of past states ${X[t-k], ..., X[t]}$, the objective is to predict future crowd states over a time horizon $\tau$. The trained model approximates an unknown transition function $f$, mapping the previous $k$ steps to the next $\tau$ steps 
\begin{equation}
    {\hat{X}[t+\tau],...,\hat{X}[t+1]} = f({X[t],...,X[t-k]}).
\end{equation}
\end{problem}

Given the crowd prediction model, we now formulate the robot's navigation task as a constrained optimization problem 
similarly to
\cite{kiss2021probabilistic}. The autonomous agent is modeled as a point $x_r \in \mathbb{R}^2$ moving with velocity $v_r \in \mathbb{R}^2$, navigating through a time-varying pedestrian field. 
The robot must plan its motion while minimizing its social impact on surrounding pedestrians, which is quantified through a social invasiveness metric, $\mathcal{I}_r$~\cite{kiss2021probabilistic}:
\begin{equation}
    \mathcal{I}_r(x_r) = \rho(x_r) (||\mu_v(x_r) - v_r||^2 + \sigma^2_v (x_r),
\label{invassivness}
\end{equation}
where the density function $\rho(x_r)$ represents the normalized pedestrian density at position $x_r$,
$\mu_v(x_r)$ denotes the expected velocity of pedestrians at that same position, while $\sigma^2_v (x_r)$ refers to the scalar variance of the pedestrians' velocity at position $x_r$.
This formulation evaluates trajectories through three key components: the likelihood of encounters through the density function $\rho$, the behavioral mismatch via relative velocity, and the motion predictability through variance. The invasiveness metric $\mathcal{I}_r(x_r)$ in \eqref{invassivness} combines these components to provide a continuous measure of the social cost associated with a trajectory. Specifically, trajectories through regions of high density $\rho(x_r)\gg0$ or significant velocity mismatch ($||\mu_v(x_r) - v_r|| \gg 0$) have higher cost. Conversely, paths through sparse areas ($\rho \approx 0$) or with matched speeds ($|\boldsymbol{\mu}_v - \mathbf{v}_r| \approx 0$) maintain lower invassivness. This formulation avoids binary thresholds by instead weighting all potential interactions.
With these definitions, the navigation objective is to minimize cumulative expected invasiveness along the trajectory. We formally state the navigation problem as follows:

\begin{problem}
Given the current robot position $x_{\text{curr}}$, the desired goal position $x_{goal}$, past observations of the environment \( \{X[t] \; | \; t \leq 0\} \), find a trajectory $\psi_r$ and arrival time $T \geq 0$ that minimize the total expected social invasiveness accumulated along the path, combined with a regularization term for path efficiency:

\begin{align}
\psi_{r}^{*}, T^{*} = \ &\arg\min_{\substack{\psi_{r} \in \Psi \\ T \in \mathbb{R}_{\geq 0}}} \int_{0}^{T} 
\bigg( \mathbb{E}\left[ \mathcal{I}_{r} \mid \{X[t] \; | \; t \leq 0\} \right] \notag \\
&\quad + \beta \left\| \frac{\mathrm{d} \psi_r[t]}{\mathrm{d}t} \right\| \bigg) \, \mathrm{d}t
\label{optimal_control}
\end{align}

where
\begin{align}
    \psi_{r}(0) &= \mathbf{x}_{curr}, \quad \text{and} \\
    \psi_{r}(T) &= \mathbf{x}_{goal}.
\end{align}
\end{problem}
Here, $\left\| \frac{\mathrm \psi_r[t]}{\mathrm{d}t} \right\|$ penalizes the distance traveled by the agent with constant velocity, weighted by the factor \(\beta\), which balances the trade-off between invasiveness and distance during the optimization of the trajectory. The factor \(\beta\) is chosen to be a small value ($\beta = 0.0001$) to ensure that the distance traveled has a small impact on the overall cost, while still guiding the robot through low-density areas.

\section{PROBABILISTIC CROWD PREDICTION}
We propose an approach to model the crowd as a macroscopic field over space and time, capturing properties such as density, average velocity, and velocity variance. A data-driven predictive model is used to forecast these properties based on historical observations, enabling real-time planning for robots operating in dynamic environments.
\subsection{Crowd Representation}
\label{crowd representation}

The environment is discretized into a 2D spatial grid with shape $H \times W$. At each time step $t$, the state of each cell $(i,j)$ is defined by a vector $x_{ij}[t]$ consisting of three components:
\begin{itemize}
    \item Density $\rho_{ij}[t]$,  represents the intensity of the spatial distribution of pedestrians at time $t$ in cell $(i,j)$. It can be interpreted as the expected number of pedestrians per unit area, offering a continuous measure of local congestion rather than a simple count.
    \item Mean velocity \(\mu_{v_{ij}}[t] \in \mathbb{R}^2\), denoting the average pedestrian velocity in the cell, expressed as a 2D vector over spatial dimensions (in the \(x\) and \(y\) directions). This is consistent with the first moment of the velocity distribution, \(\mu_{v_{ij}}(x) = \mathbb{E}[v_{ij} \mid x_{ij} = x]\).
    \item Velocity variance $\mathbf{\sigma}^2_{v_{ij}}[t]$, the scalar variance of the pedestrian velocity in the cell, quantifies motion uncertainty. It corresponds to the second moment of the velocity distribution, defined as \(\sigma_{v_{ij}}^2(x) = \mathrm{Var}[v_{ij} \mid x_{ij} = x] = \mathbb{E}[\|v_{ij} - \mu_{v_{ij}}(x)\|^2]\).
\end{itemize}

\begin{figure}[tb]
    \centering
    \includegraphics[angle=270, width=0.8\linewidth]{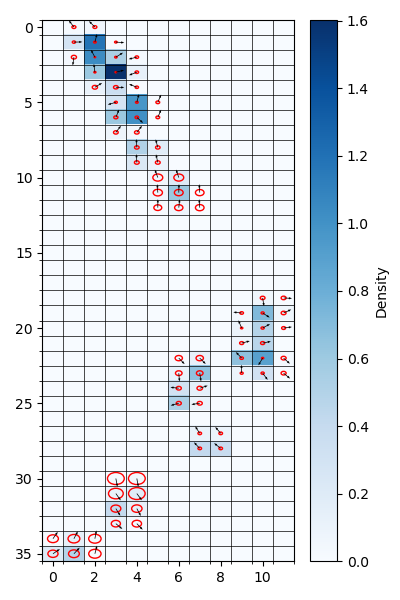}
    \caption{Visualization of a pedestrian crowd with its macroscopic characteristics. Cell colors in blue indicate the local density $\rho$, black arrows depict average velocity vectors ${\mu}_v$, red circles show the velocity variance $\sigma_v$.}
    \label{fig:corridoe_sample}
    \vspace{-2em}
\end{figure}

The global state of the environment is represented by aggregating these features across all grid cells as a 3D tensor $X[t]\subseteq\mathbb{R}^{H \ H\times W\times d} $, where \(d=4 \) is the number of features per cell, (pedestrian density $\rho$, x-velocity component $v_x$, y-velocity component $v_y$, isotropic velocity variance $\mathbf{\sigma}^2$). Fig.~\ref{fig:corridoe_sample} shows a sample of this representation.

\subsection{Spatiotemporal Prediction Model}
Based on the crowd representation introduced in Section~\ref{crowd representation}, we design a spatiotemporal forecasting architecture that models the evolution of crowd flow features over time. Spatio-temporal learning is a non-trivial challenge due to the complexities of high dimensionality. It requires capturing both spatial correlations and temporal dynamics to accurately forecast the future. 
Inspired by the work of  \cite{kiss2021probabilistic}, we employ an encoder-forecaster architecture built with ConvRNN blocks, but specifically adapted to address the unique characteristics of pedestrian dynamics.

The core of the model consists of a convolutional encoder-forecaster architecture, where the input is a sequence of crowd states over time, $X_{t-k:t}$, and the output is probabilistic predictions, $\hat X_{t+1:t+\tau}$, for $\tau$ future timestamps,
\begin{equation}
    \hat X_{t+1:t+\tau} = f_\theta (X_{t-k:t}).
\end{equation}

\begin{figure}
    \centering
    \includegraphics[width=1\linewidth]{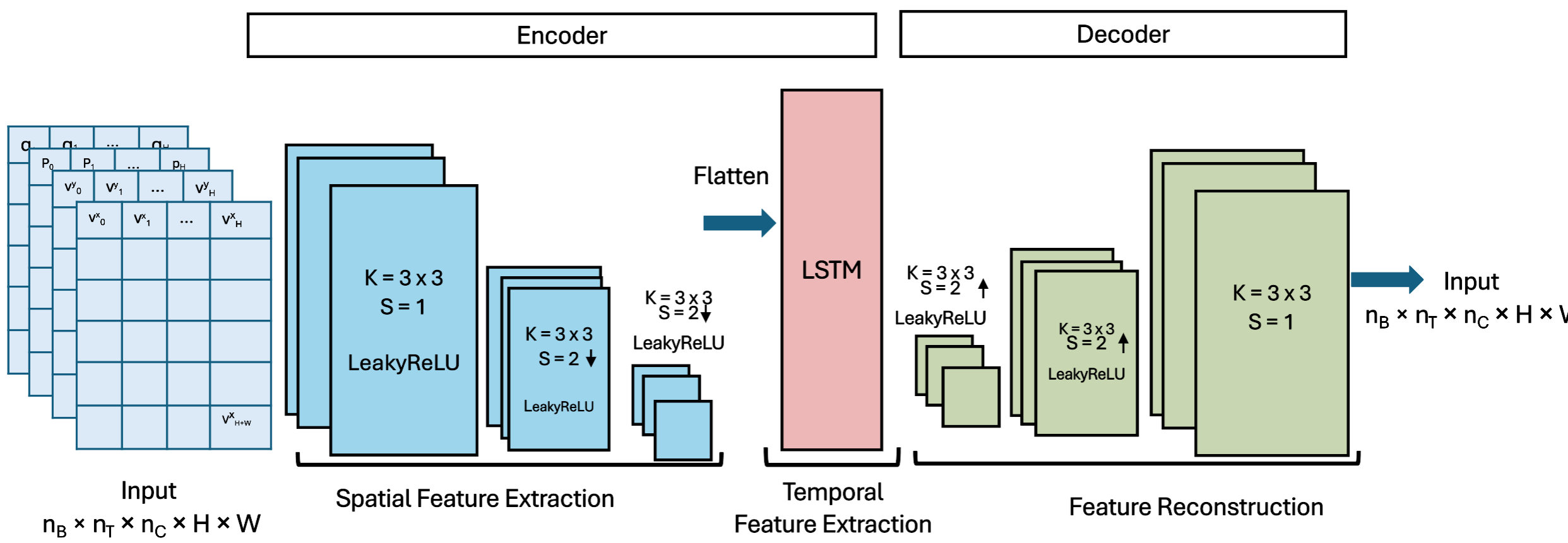}
    \caption{The proposed model processes crowd movement data through a convolutional encoder that extracts spatial features, followed by an LSTM layer that captures temporal evolution. The forecaster reconstructs predictions through transposed convolutional upsampling. Input tensors are shaped as [batch size ($n_B$), timesteps($n_T$), channels ($n_C$), grid size ($H \times W$)].}
    \label{fig:model}
    \vspace{-1.5em}
\end{figure}

The encoder consists of two convolutional layers with stride-2 reducing spatial resolution from the original $H \times W$ to $H/4 \times W/4$ while expanding channel depth from 16 to 64, thereby extracting high-level features of the crowd's spatiotemporal dynamics. 
A single \acrshort{convrnn} layer then captures temporal dynamics. This efficient design choice reflects the fundamental nature of pedestrian motion. Unlike systems requiring complex temporal processing, human crowd movement exhibits smooth velocity changes that can be effectively modeled with minimal recurrence. This recurrent layer maintains a hidden state that encodes evolving crowd dynamics and their influence on future states. For spatial reconstruction, the decoder mirrors the encoder in reverse; it upsamples the \acrshort{convrnn} output through transposed convolutions and projects it back to the original spatial resolution.  Fig.~\ref{fig:model} shows the structure of the model.

An innovative aspect of our approach lies in the design of the loss function. Rather than treating all grid cells equally, we weight the prediction errors by local pedestrian density. This density-weighted smooth L1 loss emphasizes accuracy in regions with more pedestrians while being more tolerant to errors in sparse areas. Formally, for predicted state $\hat X $ and ground truth $X$, the loss is
\begin{align}
L = \frac{1}{|V|} \Bigg[
& \sum_{(i,j) \in V} w_{\rho} \cdot \left\| \hat{\rho}_{ij} - \rho_{ij} \right\|_{L_1} \nonumber \\
& + \sum_{f \in F'} \sum_{(i,j) \in V} \rho_{ij} \cdot w_f \cdot \left\| \hat{X}_{ij}^f - X_{ij}^f \right\|_{L_1}
\Bigg]
\label{eq:loss}
\end{align}
where:
\begin{itemize}
    \item $\mathcal{V}$ denotes the set of non-empty grid cells ($\rho_{ij} > 0$);
    \item $\mathcal{F} = \{v_x, v_y, \sigma_v\}$ represents the predicted features (velocity components, and variance);
    \item $w_f$ and $w_{\rho}$ are learnable feature weights. 

\end{itemize}

In implementation, we employ a smooth L1 (Huber) loss to ensure stable gradients during training. This formulation provides several advantages over the conventional MSE loss; the density weighting $\rho_{ij}$ emphasizes accuracy in crowded regions, while the smooth L1 norm offers robustness to outliers in velocity predictions. The feature weights $w_f$ allow the model to automatically balance the relative importance of different predicted quantities during training. During implementation, we normalize the loss by the number of non-empty cells $|\mathcal{V}|$ to maintain consistent gradient scales across varying crowd densities.
\section{Socially aware motion planning}
The navigation system solves the optimization problem defined in \eqref{optimal_control} using predicted crowd states as a time-varying cost map. To enable real-time and reliable planning, we utilize the proposed model to generate future crowd dynamics and integrate this prediction capability into a graph-based planner to generate socially compliant and efficient robot trajectories introduced in~\cite{kiss2021probabilistic}. We adopt the asymptotically-optimal PRM* (Probabilistic Roadmap*)~\cite{karaman2011sampling} framework.

From the robot’s current position, we randomly sample nodes in space and time and connect forward-time neighbors to construct a spatiotemporal planning graph. 
Let $\mathbf{p}_i$ denote the spatial position of node $n_i$. The total cost of an edge connecting nodes $n_i$ and $n_j$ is defined as:
\begin{align}
    c(n_i , n_j) &= \mathcal{I}_r (n_i \rightarrow n_j) + \beta \nonumber \| \mathbf{p}_i - \mathbf{p}_j \|.
\end{align}
Here, $\mathcal{I}_r$ between two nodes is computed based on \eqref{invassivness}, quantifying how much the edge interferes with the predicted crowd flow, and the distance penalty encourages spatially efficient paths and helps prevent the planner from generating unnecessarily long or circuitous trajectories, particularly in low-density areas. This augmented cost structure ensures that the robot favors both socially aware and efficient paths.

We use Dijkstra’s algorithm~\cite{dijkstra1959195} over this graph to find the minimum-cost path to the goal and cost. To evaluate the cost, we assume the robot moves along each edge in a straight line at a constant speed. Instead of calculating the cost for every edge ahead of time, we compute each when reached. This on-demand approach makes the planning process more efficient and ensures it uses the most recent crowd predictions.

\section{Experiment and Evaluation}
Our experiments were conducted on an MSI Raider GE78HX 13VH laptop running Ubuntu 22.04.5 LTS, equipped with an Intel Core i9-13980HX processor (24 cores) @ 2.6 GHz base frequency.
\subsection{Data}
We evaluate both state estimation and planning components of our approach, using the ATC dataset~\cite{brvsvcic2013person}, which captures real-world pedestrian behavior in a large shopping mall. The dataset was collected every Wednesday and Sunday from October 24, 2012, to November 29, 2013. Since we aim to model dense social interactions, we focus exclusively on data collected on Sundays, which consistently exhibit higher pedestrian density. We also restrict the analysis to the eastern corridor. We used 28 days to train our prediction model, 9 days for validation, and 1 day for planning evaluation.

\subsection{Prediction Results}

\begin{figure}[tb]
    \centering
    \includegraphics[width=1\linewidth]{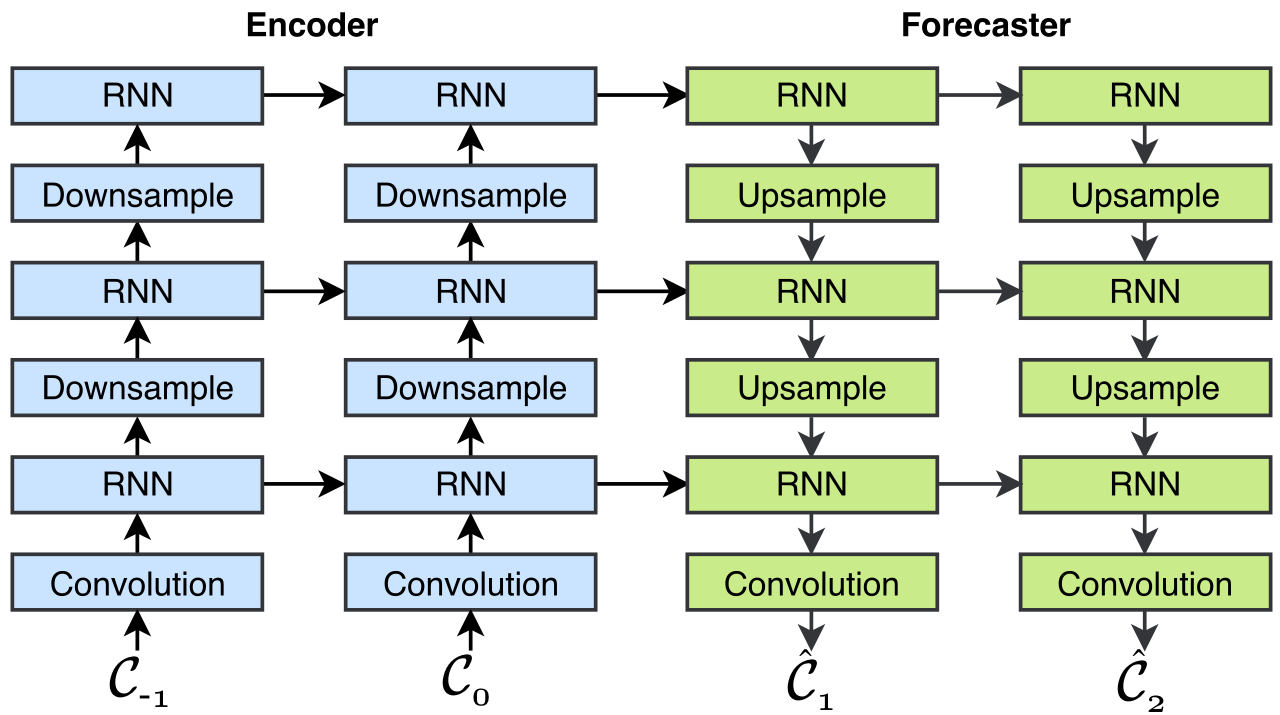}
    \caption{The encoding-forecasting framework, which predict two future frames $\hat{\mathcal{C}_1}, \hat{\mathcal{C}_2}$ based on two input frames, $\mathcal{C}_0, \mathcal{C}_{-1}$. This flexible structure accommodates varying input and output lengths. Adapted from~\cite{shi2017deep} }
    \label{fig:baseline_model}
    \vspace{-1em}
\end{figure}

\begin{figure}[tb]
    \centering
    \includegraphics[width=0.49\linewidth]{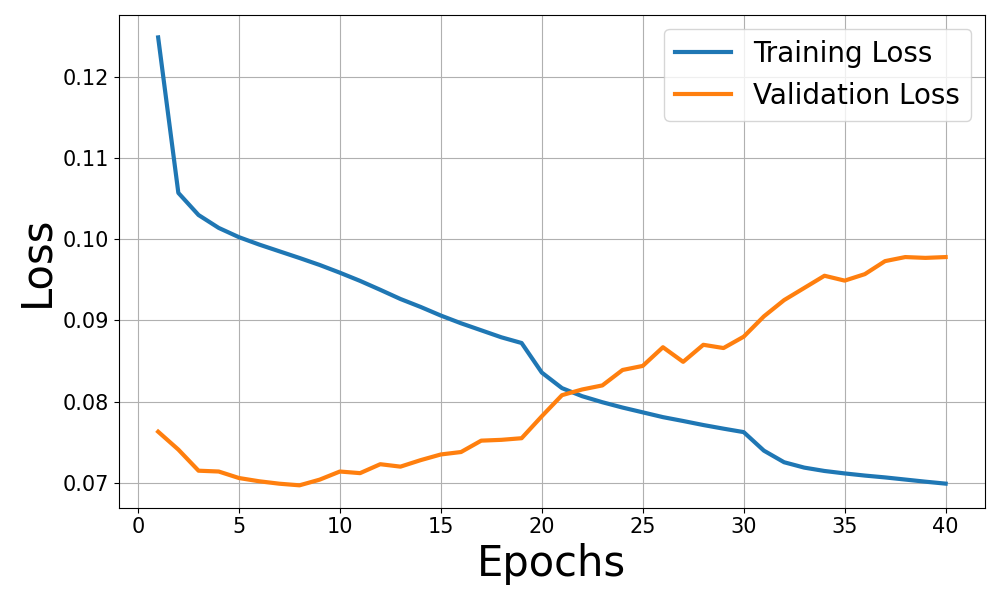}
    \includegraphics[width=0.49\linewidth]{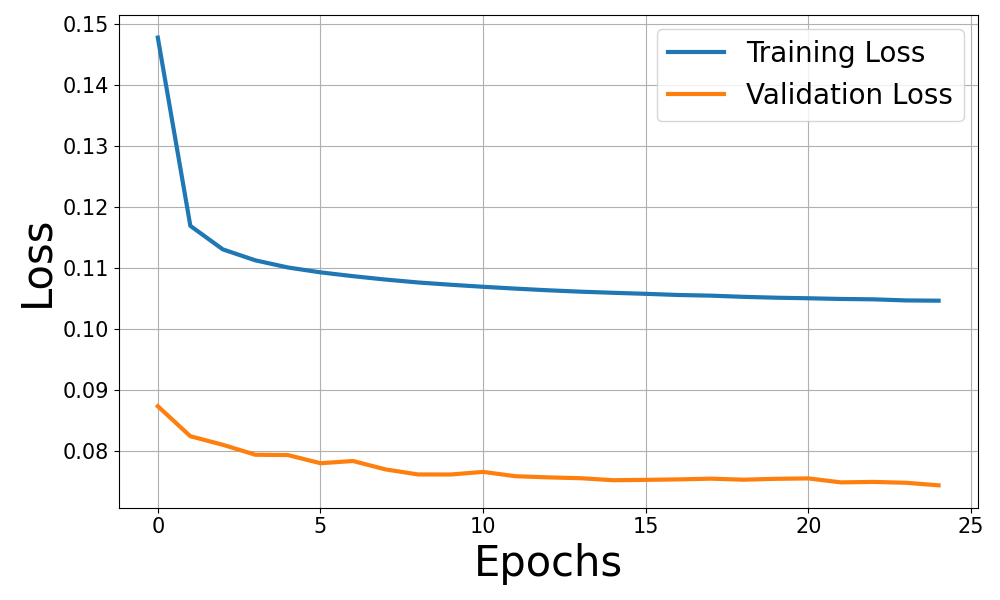}
    \caption{Learning curve comparison between (left) the baseline high-dimensional model showing overfitting tendencies and (right) the proposed lightweight model demonstrating stable generalization performance. }
    \label{fig:learning_curve}
    \vspace{-2em}
\end{figure}

To evaluate the predictive performance of propose architecture, we first convert the point-based trajectory data into a grid-based representation suitable for spatiotemporal prediction. Each observation frame is mapped into a $36 \times 12$ grid where each cell is of size $1 \si{\meter} \times 1 \si{\meter}$, and each frame corresponds to a \SI{1}{\second} time interval. For every prediction cycle, the model observes 10 frames ($k = 10$) and forecasts the subsequent 10 frames ($\tau = 10$).

We compare our lightweight architecture with a baseline model proposed in ~\cite{kiss2021probabilistic}. This model was originally developed for precipitation prediction, where chaotic, large-scale atmospheric processes demand deep, recurrent architectures. The comparative analysis between the lightweight model and the high-dimensional baseline shows that the baseline exhibits notable overfitting (see Fig.~\ref{fig:learning_curve}), while the proposed architecture maintains strong generalization throughout training.  This highlights the advantage of our approach, which avoids the complexity inherent in the baseline model. Furthermore, the lightweight model offers a substantial computational efficiency gain, achieving a 3.6-fold reduction in inference time. The baseline model requires an average of \(0.0747 \pm 0.0160 \, \si{\second}\) per prediction, while the proposed method takes only \(0.0207 \pm 0.0063 \si{\second}\).

In terms of prediction performance, Table~\ref {table:comparing_prediction} presents a comparative analysis of the baseline model and the proposed model, focusing on the density-weighted \acrfull{mae} for density, velocity, and variance. The density-weighted approach ensures particular attention to accuracy in areas of high concentration of pedestrians, where prediction quality is most operationally relevant. These results highlight that the proposed lightweight model matches, and slightly outperforms, the baseline model in all three metrics while offering substantial computational efficiency gains.

Although the baseline model is significantly more complex, it does not yield better results. Fig.~\ref{fig:baseline_model} illustrates the architecture of the original precipitation model. This design reflects the high complexity of atmospheric dynamics, using deep \acrshort{convrnn} stacks. It assumes the need for deep temporal processing and complex spatial hierarchies, which are important in capturing the chaotic nature of atmospheric systems. However, this complexity offers no advantage for crowd modeling because it is tailored to capture phenomena that are not present in pedestrian movement. Pedestrian motion is inherently smoother and more structured than atmospheric systems. While weather dynamics require deep recurrent layers to capture abrupt, chaotic changes, human crowds adjust their velocities gradually, allowing us to model temporal evolution effectively with just a single \acrshort{convrnn} layer. Similarly, the baseline’s deep spatial processing accounts for multi-scale weather interactions influenced by terrain and altitude, whereas pedestrian motion follows locally consistent patterns. Collision avoidance and flow-following behaviors remain translation-invariant across environments, eliminating the need for excessive spatial hierarchies. The baseline model's extensive feature channels, while suitable for complex atmospheric data, introduce significant overfitting risks when applied to limited pedestrian datasets.

\begin{table}[tb]
\centering
\caption{Prediction performance comparison.} 
\label{table:comparing_prediction} 
\begin{tabular}{@{}lccc@{}} 
\toprule 
\textbf{Model} & \textbf{Density MAE} & \textbf{Velocity MAE} & \textbf{Variance MAE} \\ \midrule
Baseline            &0.3978 ± 0.2120 &1.0495 ± 0.59590&0.1035 ± 0.0666 \\  
Proposed  &\textbf{0.3714 ± 0.1967} & \textbf{1.0227 ± 0.5741}&\textbf{0.0948 ± 0.0627}\\  

\bottomrule 
\end{tabular} 
\vspace{-2em}
\end{table}

\begin{figure*}[bt]
    \centering
    \includegraphics[width=0.8\linewidth]{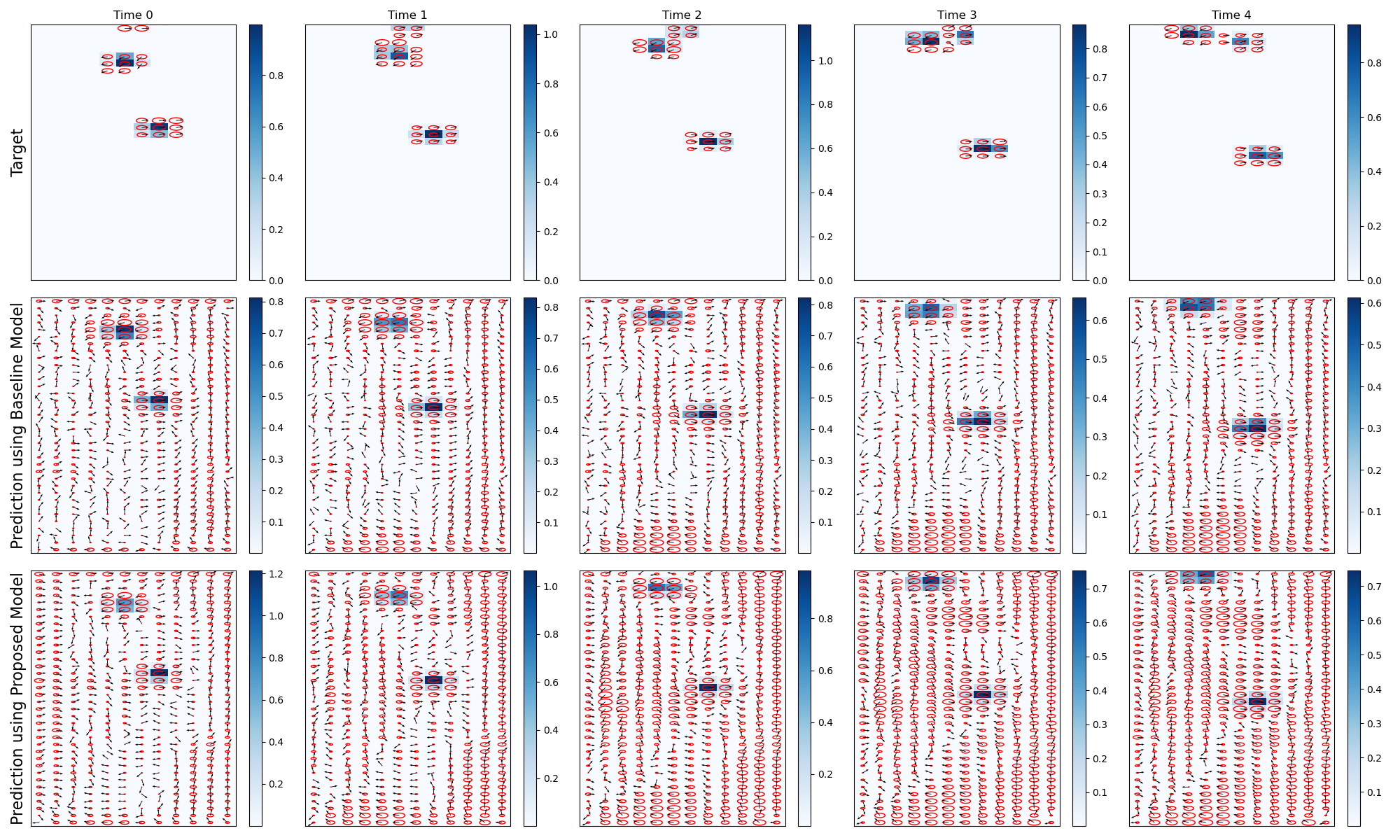}
    \caption{Macroscopic crowd prediction. From top to bottom, it shows the 5-frame ground-truth future sequence $C_{1:5}$, predictions generated by the baseline approach~\cite{karaman2011sampling}, and the proposed model $\hat{C}_{1:5}$. The crowd is represented as described in the caption of Fig.~\ref{fig:corridoe_sample}.}
    \label{fig:prediction_sample}
    \vspace{-1.5em}
\end{figure*}

Finally, Fig.~\ref{fig:prediction_sample} provides a visual example comparing crowd predictions by the baseline and the proposed model. The visual comparison further illustrates that the proposed model, while simplified, can effectively capture the key dynamics of crowd movement for real-time planning applications.

\subsection{Planning Evaluation}
Using environment-specific crowd predictions (trained on the ATC dataset), we implement a trajectory planner that integrates the proposed model within a spatio-temporal PRM* framework, following the structure introduced in~\cite{kiss2021probabilistic}. Sampled nodes represent positions and times, and are connected forward in time.
Each edge in the roadmap is evaluated based on its invasiveness cost. In contrast to the baseline, we augment the invasiveness metric with a distance-based penalty to discourage unnecessarily long trajectories in low-density areas.

We compare our planner against four predictive models. We adopt two settings from the baseline method; the offline and the online models. In the offline setting, the prediction model observes a short sequence of past crowd states and produces a single forecast, which is then used to plan the entire trajectory without further updates. In the online variant, the planner continuously updates the prediction at regular intervals, re-planning based on the latest observations of the environment. For upper-bound comparison, we include an omniscient model, which assumes perfect knowledge of future crowd states. Although not achievable in practice, it provides a reference for the best possible performance under ideal conditions. Finally, we introduce our model, which allows for rapid inference into the planning, enabling fast, socially compliant trajectory generation in real-time.

We evaluate the performance of each planning strategy by computing both expected and actual trajectory costs over a set of test scenarios, as defined in the optimization objective of Equation~\eqref{optimal_control}. The expected cost is calculated based on predicted crowd states, while the actual cost is computed using the true crowd motion. Table~\ref {table:planning_cost} presents the comparative results, aggregating data from $16$ distinct test scenarios across $5$ different time windows to capture varying crowd density conditions. The reported values represent averages over $80$ planning instances per method.

\begin{figure}[tb]
    \centering
    \includegraphics[angle=0, width=1\linewidth]{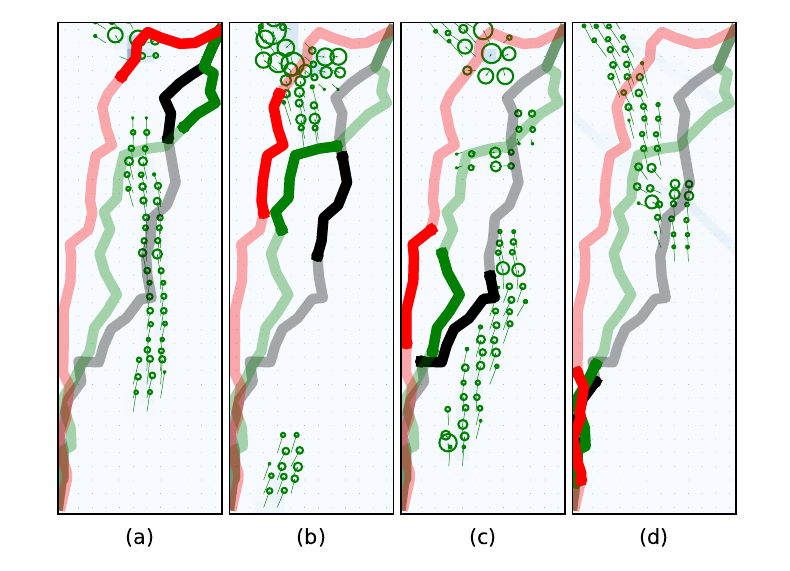}
    \caption{Trajectory planning comparison across methods. From right to left, each frame shows 10-second increments. Colored paths represent: baseline online (red), proposed online (green), and omniscient reference (black). Solid segments indicate current positions, while translucent lines show complete trajectories. The background shows the crowd representation (see Fig. \ref{fig:corridoe_sample}). In Fig. (a), the baseline trajectory (red) interacts with the crowd, resulting in a higher traversal cost, whereas the proposed (green) and omniscient (black) trajectories effectively avoid the crowd. A similar behavior is observed in Fig. (b). In Fig. (c), all trajectories successfully avoid the crowd, and in Fig. (d), all trajectories have reached the destination.}
    \label{fig:planned_trajectories}
\end{figure}

The experimental results demonstrate that our proposed planner achieves marginally lower actual trajectory costs, and significantly more accurate cost estimation during planning ($30.3\%$ reduction in estimating expected cost). The strong alignment between expected and actual costs confirms our model's enhanced forecasting capability, suggesting more reliable planning decisions in practice.
Fig.~\ref{fig:planned_trajectories} shows trajectories generated with the baseline model, proposed model (online planning) and with the ground truth in one scenario. 

\begin{table}[tb]
\centering
\caption{Planning performance comparison across methods} 
\label{table:planning_cost} 
\begin{tabular}{@{}lcc@{}} 
\toprule 
Method               & Expected Cost & Actual Cost \\ \midrule 
Baseline (once)      & $0.1247 \pm 0.0186$    & $0.0619 \pm 0.0080$      \\  
Baseline (online)    & $0.0798 \pm 0.0156$    & $0.0518 \pm 0.0084$     \\  
Proposed (once)      & $0.1076 \pm 0.0138$    & $0.0524 \pm 0.0068$     \\  
Proposed (online)    & $\mathbf{0.0556 \pm 0.0086}$ & $\mathbf{0.0502 \pm 0.0053}$     \\  
Omniscient          &                         & $0.0347 \pm 0.0040$     \\  
\bottomrule 
\end{tabular} 
\vspace{-2.5em}
\end{table}


\section{Conclusion}
This paper presented a lightweight yet effective approach for crowd modeling, and we presents its applicability to crowd-aware robot navigation.
The proposed method successfully balances prediction accuracy with computational efficiency. By fundamentally redesigning a weather-inspired architecture to match the unique characteristics of pedestrian motion, our model achieves improved generalization and prediction performance, and faster inference time. The planning experiments demonstrate that our simplified model not only matches but slightly outperforms the baseline in actual trajectory costs while providing significantly more accurate cost estimates during planning. This work establishes that effective crowd navigation 
can be achieved with substantially smaller 
carefully tailored models for human motion patterns.

\addtolength{\textheight}{-12cm}   





\bibliography{ref}  

\begin{thebibliography}{10}
\providecommand{\url}[1]{#1}
\csname url@samestyle\endcsname
\providecommand{\newblock}{\relax}
\providecommand{\bibinfo}[2]{#2}
\providecommand{\BIBentrySTDinterwordspacing}{\spaceskip=0pt\relax}
\providecommand{\BIBentryALTinterwordstretchfactor}{4}
\providecommand{\BIBentryALTinterwordspacing}{\spaceskip=\fontdimen2\font plus
\BIBentryALTinterwordstretchfactor\fontdimen3\font minus \fontdimen4\font\relax}
\providecommand{\BIBforeignlanguage}[2]{{%
\expandafter\ifx\csname l@#1\endcsname\relax
\typeout{** WARNING: IEEEtran.bst: No hyphenation pattern has been}%
\typeout{** loaded for the language `#1'. Using the pattern for}%
\typeout{** the default language instead.}%
\else
\language=\csname l@#1\endcsname
\fi
#2}}
\providecommand{\BIBdecl}{\relax}
\BIBdecl

\bibitem{triebel2016spencer}
R.~Triebel, K.~Arras, R.~Alami, L.~Beyer, S.~Breuers, R.~Chatila, M.~Chetouani, D.~Cremers, V.~Evers, M.~Fiore \emph{et~al.}, ``Spencer: A socially aware service robot for passenger guidance and help in busy airports,'' in \emph{Field and Service Robotics: Results of the 10th International Conference}.\hskip 1em plus 0.5em minus 0.4em\relax Springer, 2016, pp. 607--622.

\bibitem{sabelli2016robovie}
A.~M. Sabelli and T.~Kanda, ``Robovie as a mascot: a qualitative study for long-term presence of robots in a shopping mall,'' \emph{International Journal of Social Robotics}, vol.~8, pp. 211--221, 2016.

\bibitem{singamaneni2024survey}
P.~T. Singamaneni, P.~Bachiller-Burgos, L.~J. Manso, A.~Garrell, A.~Sanfeliu, A.~Spalanzani, and R.~Alami, ``A survey on socially aware robot navigation: Taxonomy and future challenges,'' \emph{The International Journal of Robotics Research}, vol.~43, no.~10, pp. 1533--1572, 2024.

\bibitem{mavrogiannis2023core}
C.~Mavrogiannis, F.~Baldini, A.~Wang, D.~Zhao, P.~Trautman, A.~Steinfeld, and J.~Oh, ``Core challenges of social robot navigation: A survey,'' \emph{ACM Transactions on Human-Robot Interaction}, vol.~12, no.~3, pp. 1--39, 2023.

\bibitem{molina2019go}
S.~Molina, G.~Cielniak, and T.~Duckett, ``Go with the flow: Exploration and mapping of pedestrian flow patterns from partial observations,'' in \emph{International Conference on Robotics and Automation (ICRA)}.\hskip 1em plus 0.5em minus 0.4em\relax IEEE, 2019, pp. 9725--9731.

\bibitem{liu2025improved}
A.~Liu, C.~Liu, L.~Li, R.~Wang, and Z.~Lu, ``An improved fuzzy-controlled local path planning algorithm based on dynamic window approach,'' \emph{Journal of Field Robotics}, vol.~42, no.~2, pp. 430--454, 2025.

\bibitem{mehran2009abnormal}
R.~Mehran, A.~Oyama, and M.~Shah, ``Abnormal crowd behavior detection using social force model,'' in \emph{IEEE conference on computer vision and pattern recognition}, 2009, pp. 935--942.

\bibitem{cheng2024multi}
C.-L. Cheng, C.-C. Hsu, S.~Saeedvand, and J.-H. Jo, ``Multi-objective crowd-aware robot navigation system using deep reinforcement learning,'' \emph{Applied Soft Computing}, vol. 151, p. 111154, 2024.

\bibitem{trautman2010unfreezing}
P.~Trautman and A.~Krause, ``Unfreezing the robot: Navigation in dense, interacting crowds,'' in \emph{IEEE/RSJ International Conference on Intelligent Robots and Systems}, 2010, pp. 797--803.

\bibitem{alahi2016social}
A.~Alahi, K.~Goel, V.~Ramanathan, A.~Robicquet, L.~Fei-Fei, and S.~Savarese, ``Social {LSTM}: Human trajectory prediction in crowded spaces,'' in \emph{Proceedings of the IEEE conference on computer vision and pattern recognition}, 2016, pp. 961--971.

\bibitem{gupta2018social}
A.~Gupta, J.~Johnson, L.~Fei-Fei, S.~Savarese, and A.~Alahi, ``Social {GAN}: Socially acceptable trajectories with generative adversarial networks,'' in \emph{Proceedings of the IEEE conference on computer vision and pattern recognition}, 2018, pp. 2255--2264.

\bibitem{sadeghian2019sophie}
A.~Sadeghian, V.~Kosaraju, A.~Sadeghian, N.~Hirose, H.~Rezatofighi, and S.~Savarese, ``Sophie: An attentive {GAN} for predicting paths compliant to social and physical constraints,'' in \emph{Proceedings of the IEEE/CVF conference on computer vision and pattern recognition}, 2019, pp. 1349--1358.

\bibitem{vemula2018social}
A.~Vemula, K.~Muelling, and J.~Oh, ``Social attention: Modeling attention in human crowds,'' in \emph{international Conference on Robotics and Automation (ICRA)}.\hskip 1em plus 0.5em minus 0.4em\relax IEEE, 2018, pp. 4601--4607.

\bibitem{huang2025interaction}
Z.~Huang, T.~Ji, H.~Zhang, F.~C. Pouria, K.~Driggs-Campbell, and R.~Dong, ``Interaction-aware conformal prediction for crowd navigation,'' \emph{arXiv preprint arXiv:2502.06221}, 2025.

\bibitem{treuille2006continuum}
A.~Treuille, S.~Cooper, and Z.~Popovi{\'c}, ``Continuum crowds,'' \emph{ACM transactions on graphics (TOG)}, vol.~25, no.~3, pp. 1160--1168, 2006.

\bibitem{narain2009aggregate}
R.~Narain, A.~Golas, S.~Curtis, and M.~C. Lin, ``Aggregate dynamics for dense crowd simulation,'' in \emph{ACM SIGGRAPH Asia 2009 papers}, pp. 1--8.

\bibitem{kiss2021probabilistic}
S.~H. Kiss, K.~Katuwandeniya, A.~Alempijevic, and T.~Vidal-Calleja, ``Probabilistic dynamic crowd prediction for social navigation,'' in \emph{international conference on robotics and automation (ICRA)}.\hskip 1em plus 0.5em minus 0.4em\relax IEEE, 2021, pp. 9269--9275.

\bibitem{shi2015convolutional}
X.~Shi, Z.~Chen, H.~Wang, D.-Y. Yeung, W.-K. Wong, and W.-c. Woo, ``Convolutional {LSTM} network: A machine learning approach for precipitation nowcasting,'' \emph{Advances in neural information processing systems}, vol.~28, 2015.

\bibitem{karaman2011sampling}
S.~Karaman and E.~Frazzoli, ``Sampling-based algorithms for optimal motion planning,'' \emph{The international journal of robotics research}, vol.~30, no.~7, pp. 846--894, 2011.

\bibitem{dijkstra1959195}
E.~Dijkstra, ``A note on two problems in connection with graphs,'' \emph{Numerische Mathematik}, pp. 1--269, 1959.

\bibitem{brvsvcic2013person}
D.~Br{\v{s}}{\v{c}}i{\'c}, T.~Kanda, T.~Ikeda, and T.~Miyashita, ``Person tracking in large public spaces using 3-d range sensors,'' \emph{Transactions on Human-Machine Systems}, vol.~43, no.~6, pp. 522--534, 2013.

\bibitem{shi2017deep}
X.~Shi, Z.~Gao, L.~Lausen, H.~Wang, D.-Y. Yeung, W.-k. Wong, and W.-c. Woo, ``Deep learning for precipitation nowcasting: A benchmark and a new model,'' \emph{Advances in neural information processing systems}, vol.~30, 2017.

\end{thebibliography}

\end{document}